\documentclass{article}
\usepackage{ijcai22}

\usepackage{times}

\usepackage{soul}
\usepackage{url}
\usepackage[hidelinks]{hyperref}
\usepackage[utf8]{inputenc}
\usepackage[small]{caption}
\usepackage{graphicx}
\usepackage{amsmath}
\usepackage{booktabs}
\usepackage{natbib}

\title{Training Naturalized Semantic Parsers with Very Little Data}

\author{
Subendhu Rongali$^{1,2}$\footnote{Work done during the author's summer internship.}\and
Konstantine Arkoudas$^2$\and
Melanie Rubino$^2$\and
Wael Hamza$^2$\\
\affiliations
$^1$University of Massachusetts Amherst\\
$^2$Amazon Alexa AI, New York\\
\emails
srongali@cs.umass.edu,
\{arkoudk, rubinome, waelhamz\}@amazon.com
}

\usepackage[switch]{lineno}

\begin{document}

\maketitle


\begin{abstract}

Semantic parsing is an important NLP problem, particularly for voice assistants such as Alexa and Google Assistant. State-of-the-art (SOTA) semantic parsers are seq2seq architectures based on large language models that have been pretrained on vast amounts of text. To better leverage that pretraining, recent work has explored a reformulation of semantic parsing whereby the output sequences are themselves natural language sentences, but in a controlled fragment of natural language. This approach delivers strong results, particularly for few-shot semantic parsing, which is of key importance in practice and the focus of our paper. We push this line of work forward by introducing an automated methodology that delivers very significant additional improvements by utilizing modest amounts of unannotated data, which is typically easy to obtain. Our method is based on a novel synthesis of four techniques: joint training with auxiliary unsupervised tasks; constrained decoding; self-training; and paraphrasing. We show that this method delivers new SOTA few-shot performance on the Overnight dataset, particularly in very low-resource settings, and very compelling few-shot results on a new semantic parsing dataset.

\end{abstract}

\section{Introduction}

Semantic parsing is the task of mapping a natural-language utterance to a structured representation of the meaning of the utterance.  Often, the output meaning representation is a formula in an artificial language such as SQL or some type of formal logic. Current SOTA semantic parsers are seq2seq architectures based on very large language models (LMs) that have been pretrained on vast amounts of natural-language text \citep{rongali2020don,einolghozati2019improving}. To better capitalize on that pretraining, various researchers have proposed to reformulate semantic parsing so that the output meaning representation is itself expressed in natural---instead of a formal---language, albeit a controlled (or ``canonical'') fragment of natural language that can then be readily parsed into a conventional logical form (LF). We refer to this reformulation as the {\em naturalization\/} of semantic parsing.

Naturalizing a semantic parser has significant advantages because the reformulated task involves natural language on both the input and the output space, making it better aligned with the pretraining LM objective. However, even with large-scale LM pretraining, fine-tuning these models requires lots of data, and producing complex annotations for semantic parsing is expensive. There has hence been great interest in {\em few-shot\/} semantic parsing, where we only have access to a few annotated examples \citep{shin2021constrained,xu2020autoqa}.

Techniques such as in-context learning and prompting \citep{shin2021constrained} 
have shown very promising results in few-shot scenarios for semantic parsing when used with extremely large (and access-restricted) pretrained LMs such as GPT-3 (175B parameters). However, the size and inaccessibility of these models makes their use infeasible at present. Task-specific fine-tuning of smaller LMs remains the best-performing approach that is practically feasible, and is the one we pursue in this paper. We propose a simple but highly effective methodology for few-shot training of naturalized semantic parsers that can be used with smaller and more ecologically friendly LMs (we use BART-Large, which has fewer than 0.5B parameters) and can be quickly applied to bootstrap a high-performing semantic parser with less than 50 annotated examples and a modest number of unlabeled examples, which are typically readily available (and can often be readily synthesized).

Our methodology is based on a judicious composition of four techniques: joint training of the semantic parsing task with masking and denoising LM objectives; constrained decoding, made possible because the canonical fragment of natural language is generated by a simple grammar; self-training; and paraphrasing.
For training dataset sizes ranging from $n = 16$ to $n = 200$, our method consistently outperforms previous BART-based and GPT-2-based few-shot SOTA
results on all domains of the Overnight dataset, in some cases delivering relative improvements exceeding 100\%. For $n = 200$, our method catches
up to and slightly outperforms in-context learning with GPT-3. We also provide results on Pizza, a new semantic parsing dataset, where
we demonstrate relative improvements over BART-based SOTA architectures ranging from 20\% to 190\%. 
\begin{figure}[t!]
  \centering
  \includegraphics[width=\linewidth]{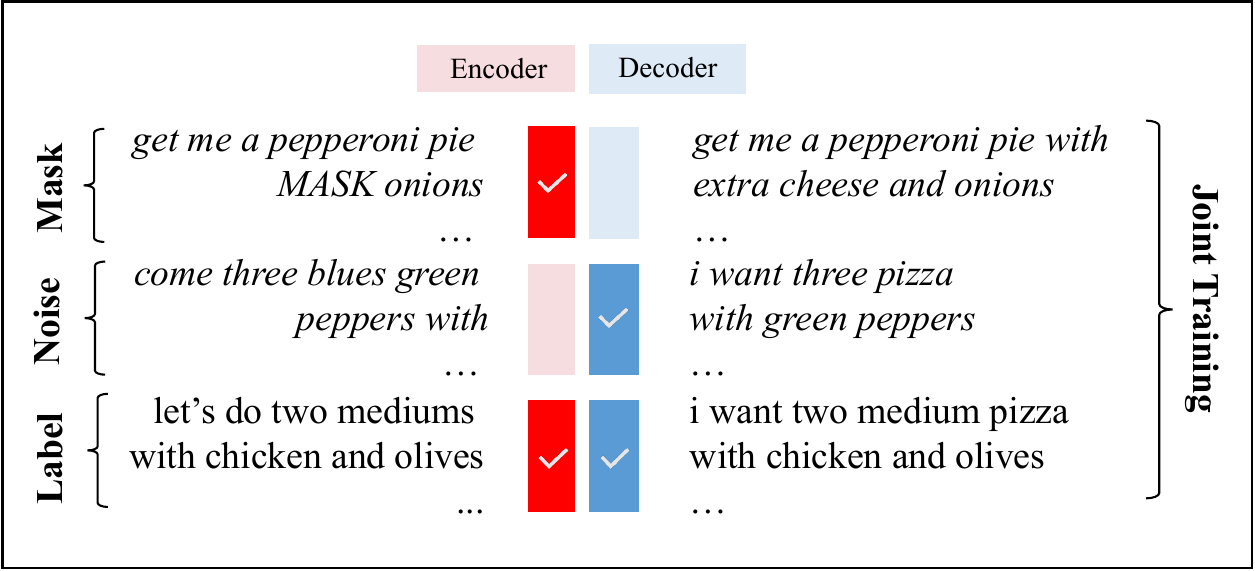}
  \caption{Jointly training a seq2seq model using mask prediction, denoising, and supervised semantic parsing examples.}
  \label{fig:mf}
\end{figure} 

We start with the best-performing finetuned naturalized model from \citet{shin2021constrained} as our baseline.
This model is based on BART \citep{lewis2019bart}, which was chosen by the authors because both the encoder and the decoder are pretrained. They also constrain their decoder to produce only
  valid canonical forms, 
  using a method that filters valid next tokens. The authors showed that these techniques greatly improve model robustness and allowed
  models to train with just a few hundred examples.
  
  We pursue the same general direction here but propose a general methodology that leverages modest amounts of unannotated data 
  to deliver very significant improvements over that baseline model without needing additional effort from model developers. 
  Specifically, we use unlabeled user utterances to create a masked prediction task, which allows the encoder to see and learn to encode
  utterances of interest. We then add random noise to a generated target dataset to produce noisy source sequences and create an additional
  denoising task. This task trains the decoder to produce canonical forms effectively. We merge the source and target sequences from
  both of these tasks along with the original labeled set and train a BART model. Figure~\ref{fig:mf} illustrates this process. 
  At inference, we use constrained decoding, which ensures that we only generate valid canonical forms. By augmenting the dataset with
  additional examples that effectively adapt the encoder and decoder, we observe massive improvements in semantic parsing accuracy
  over the baseline models that are only fine-tuned on the labeled dataset.
 Apart from joint training (JT for short), our method uses self-training \citep{mcclosky2006effective,goldwasser2011confidence}, or ST for short,
 and paraphrase augmentation \citep{feng2021survey,xu2020autoqa}. Here, we take the model from the JT step and label all
  the unlabeled utterances with constrained decoding. We also paraphrase all our utterances to create more data and label them
  in the same way. We then repeat the JT step with this enlarged self-labeled dataset, the original golden labeled dataset,
  and the masked and noised datasets. Since the self-labeling is done in a constrained manner, the labels are corrected if 
  our model slightly strays from the golden parses. Injecting this knowledge back into the model helps it improve even further.

\section{Methodology}

\subsection{Base model}
\label{sec:mp}
Our starting architecture is based on the best fine-tuning model reported by \citet{shin2021constrained}, which is a BART-Large \citep{lewis2019bart} seq2seq model
with canonical-form targets and constrained decoding.
Since this architecture uses canonical-form targets, both the inputs and the outputs are English sentences. As an example from the Pizza dataset,
an input utterance like \emph{could i have a medium pie along with ham sausage but please avoid bacon} is mapped to the output target
\emph{i want one medium pizza with ham and sausage and no bacon}. Canonical forms are defined by the domain developers and are
designed so that they can be easily parsed using simple rules to obtain a conventional target LF. 

Since the target canonical forms are user-defined and generated by a fixed grammar from which the 
ultimate meaning representations can be recovered, we can constrain the decoding in the seq2seq model to only produce
valid sequences adhering to that grammar. We do this by defining a {\tt validNextTokens}
function that takes the tokens generated so far as input and returns the valid set of next tokens. During beam
search, we adjust the logits to filter out invalid tokens.

\subsection{Joint Training}
We now describe our novel JT technique. While the base architecture was shown to perform well with dataset sizes of around 200,
we observed that there is a lot of room for improvement when the number of annotated examples falls further (to 48, 32, and 16 examples).
Our key idea was to introduce auxiliary tasks constructed from easily-obtainable unsupervised data, and jointly train the
model on these tasks, in addition to the semantic parsing task with a very small number of labeled examples.
While labeling utterances is expensive, one can assume access to a larger set of unlabeled utterances. 
This assumption holds especially true for commercial voice assistants, which can record de-identified live traffic
from participating customers. But even when bootstrapping new semantic parsers in a cold-start scenario, it is much easier
to come up with utterances that need to be supported than it is to annotate these utterances. 
We can also generate a lot of target parse trees or canonical forms automatically, by sampling and generating from
the target grammar. For example, we can generate sample pizza orders and create corresponding canonical forms.
Given such data, we construct two tasks, Mask Prediction and Denoising, to augment the regular task. 

\subsubsection{Mask Prediction}

Our first auxiliary task is focused on improving the encoder. We would like the encoder to see and learn to encode real
source utterance sequences. To accomplish this, we use the unlabeled user utterances to construct an infilling-style 
mask prediction task. We mask spans of tokens in the same style as the BART pretraining objective. 
As an example where we mask a span containing roughly 25\% of the tokens, the source is 
\emph{i'll go for five pizzas along with MASK but avoid sausage} and the target is 
\emph{i'll go for five pizzas along with mushrooms and onions but avoid sausage}. 
This task can be viewed 
as a form of domain adaptation, where the BART pretraining is continued on domain-specific data.
It hence acts as a potential regularizer that stabilizes training with a small downstream task dataset.
However, as we will show later, integrating it with the regular task via JT is more effective than
first adapting and then only fine-tuning on the labeled data. 

\subsubsection{Denoising}
Our second auxiliary task is focused on improving the decoder.
For this, we use the synthesized target canonical forms. 
These target canonical forms can be synthesized easily by randomly sampling from the target grammar.
With the pizza dataset for example, this just corresponds to randomly creating various pizza orders and constructing 
their canonical forms. With a dataset like Overnight, it corresponds to generating random database queries from the query grammar.

Once we have a large set of random targets, we create a noisy version to use as the source
sequences for a denoising task. We only add noise to the non-content tokens, i.e., tokens
that do not interfere with entity names or intents. We do this to ensure that the model does not hallucinate. 
The choice of canonical forms which contain natural language instead of parse trees is also important here,
as it allows us to easily add such noise. The noise itself consists of a set of manipulations on tokens. We randomly choose from the
five following operations to apply to tokens with a certain probability:
\begin{itemize}
\item Delete: Delete a token.
\item Replace: Replace a token with a token randomly sampled from the vocabulary.
\item Swap: Swap two consecutive tokens.
\item Insert: Insert a randomly sampled  token.
\item Duplicate: Duplicate a token. 
\end{itemize}

An example of a noisy source sequence is \emph{dishes want pizza one notified banana peppers uty pickles},
where the target is \emph{i want one pizza with banana peppers and pickles}.
Once we construct the mask prediction and denoising datasets, we combine them with the
labeled semantic parsing examples. We then shuffle the entire dataset and train the 
BART seq2seq model. Note that we do not introduce any weights or custom loss functions. We 
simply use the original sequence prediction loss to train on the new augmented dataset, as shown in 
Figure~\ref{fig:mf}. We also do not explicitly differentiate between different task examples. The model learns to 
do mask prediction when it sees a \emph{MASK} token. If not, it tries to generate 
a canonical form target sequence. We further ensure this is the case during inference 
by using constrained decoding.


\subsection{Self-Training and Paraphrasing}
To further improve upon JT, we introduce two enhancements: self-training and paraphrase augmentation. While both 
have been previously explored in isolation, we show that they work better in tandem with JT.

Self-training is a popular semi-supervised learning technique that has been explored across a wide range of 
applications to improve models with limited annotated data \citep{mcclosky2006effective,mihalcea2004co}. 
The key idea is to first build a model with the existing labeled data and then use it to annotate an unlabeled dataset in order
to obtain noisy annotations (silver labels). The model is then retrained with the combination of 
the original golden plus the silver data. 
This approach typically works
well in low-resource scenarios for classification-style tasks or tasks with limited annotation diversity.  It also
requires a reasonable initial checkpoint to obtain the silver annotations.

We use our joint trained model as the initial checkpoint to label data. We make the self-training approach more 
effective for our generation-style task with some important design choices. The constrained decoding 
improves the label quality of the silver annotations and injects additional knowledge to retrain the model. We also 
add the mask prediction and denoising datasets to better retrain the model. Note that we do not perform any
confidence-based filtering or re-ranking on the silver labels 
since partially correct data might still help the decoder and confidence scores aren't reliable \citep{NSPConfidence}, especially with 
constrained decoding.
We simply obtain predictions for 
all unlabeled data and use them to retrain the model, making this a straightforward enhancement.

A significant improvement comes from the data diversity introduced by self-training. By labeling unannotated utterances 
and augmenting the training dataset, the retrained model sees a larger variety of utterances. 
To further increase 
this variety, we propose paraphrase augmentation.

Paraphrasing is an effective way to obtain similar sentences with different surface forms. Since most neural paraphrasing models
are noisy, especially when applied to out-of-domain data, we cannot assume that
the semantics of the paraphrases are still captured by the original golden annotations. Instead, we
rely on the self-training approach and use the JT model to label the newly generated paraphrases. We found that the diversity from the silver labeled utterances from these techniques is useful upto a certain size, at which point the label noise overpowers the diversity gains.

For our experiments, we built a paraphrasing model by training a BART-Large model on 5m examples 
from the ParaNMT dataset \citep{wieting2017paranmt} for two epochs. As an example, \emph{how many all season fouls did kobe bryant have as an la laker}
is paraphrased as \emph{how many fouls did kobe bryant have as a lakers player}. These are not exact paraphrases but still serve as 
new utterances for self-training.

\subsection{Bringing it all together}
To summarize, we start with a BART-Large seq2seq model. We convert the target LFs into canonical natural-language forms and implement
constrained decoding to ensure that the generated tokens represent valid canonical forms. This is our base architecture.
We train this base model using JT with mask infilling and denoising as additional auxiliary tasks, along with 
semantic parsing using the limited labeled data. This produces our first model.

We use this model to label any available unannotated utterances. We also paraphrase all the utterances and label the paraphrases with the same model. We then augment
the JT data with these newly labeled examples and retrain the model. We get two more models in this step,
one that uses the paraphrased data and one that does not.

\section{Experimental Setup}

\subsection{Datasets}
\label{sec:ds}
We evaluate our techniques on two datasets: Pizza\footnote{\footnotesize https://github.com/amazon-research/pizza-semantic-parsing-dataset}
and Overnight \citep{wang2015building}. We use three
low-resource  settings: 16, 32, and 48 labeled examples. These are randomly sampled from the full original datasets.

Pizza is a recently introduced dataset consisting of English utterances that represent orders of pizzas and drinks. 
The target parse is a LF that specifies the various components of the relevant
pizza and drink orders. An example from this dataset was given in Section~\ref{sec:mp}. We defined a canonicalization scheme for pizza and drink orders via a rule-based parser that can go from the canonical form to the LF and conversely. 

The original Pizza dataset contains a synthetic training set, and real dev and test sets.
For our experiments, we use the dev set to choose example for low-resource training.  For the denoising task, we randomly sample 10k 
target parses from the original synthetic training set and construct their canonical forms to simulate random pizza orders.

Overnight is a popular semantic parsing dataset that consists of 13,682 examples 
across eight domains. The task is to convert natural language utterances to database queries, which are then
executed to obtain the results for the user utterances. We have 
access to the utterance, canonical form and the corresponding database query for all examples. An example
from the basketball domain is the utterance \emph{which team did kobe bryant play on in 2004}, whose canonical
target is \emph{team of player kobe bryant whose season is 2004}.

To generate queries for the denoising task, we use the SEMPRE toolkit \citep{berant2013freebase}, upon which 
the Overnight dataset was built, to generate sample queries for each domain from its canonical grammar, 
consisting of around 100 general and 20-30 per-domain rules.

For both datasets, for paraphrase augmentation, we generate four paraphrases for each 
utterance in the training set. We use the BART-Large model trained on ParaNMT data and take the top four sequences from 
beam search decoding at inference. For constrained decoding, we follow the approach of \citet{shin2021constrained} and construct a large trie 
that contains all the canonical form sequences, and use it to look up valid next tokens given a prefix. 

\begin{table}
\centering
{\small \begin{tabular}{lccc}\toprule
& \multicolumn{3}{c}{Unordered EM Accuracy}
\\\cmidrule(lr){2-4}
           & n=16  & n=32 & n=48 \\\midrule
          \multicolumn{4}{c}{Baselines} \\\midrule
BART Canonical    & 16.95 & 53.35 & 58.36 \\
\midrule
 \multicolumn{4}{c}{Our models} \\\midrule
JT & 42.23 & \bf 64.70 & 70.30 \\
JT + Self Training & \bf 49.89 & 63.23 & 72.07 \\
JT + Self Training + Paraphrasing & 48.19 & 64.55 & \bf 73.10 \\\midrule
\multicolumn{4}{c}{Reference} \\\midrule
Full BART Canonical    & \multicolumn{3}{c}{87.25 (n = 348)}  \\
\bottomrule
\end{tabular}}
\caption{Results on the Pizza dataset.
}
\label{tab:pizza}
\end{table}

\subsection{Baseline Models}
We compare our models to the best fine-tuned model from \citet{shin2021constrained}, a BART-Large seq2seq model
with canonical-form targets and constrained decoding, which we use as our base architecture.
This model is trained on the same data as our JT models in the low-resource settings.
We also compare to a fully trained version of this model, trained on all available training data. 

\begin{table*}[t!]
\centering
{\footnotesize \begin{tabular}{lcccccccccccc}\toprule
& \multicolumn{3}{c}{Basketball}& \multicolumn{3}{c}{Calendar}& \multicolumn{3}{c}{Publications}& \multicolumn{3}{c}{Restaurants}
\\\cmidrule(lr){2-4}\cmidrule(lr){5-7}\cmidrule(lr){8-10}\cmidrule(lr){11-13}
           & n=16  & n=32 & n=48 & n=16  & n=32 & n=48 & n=16  & n=32 & n=48 & n=16  & n=32 & n=48 \\\midrule
          \multicolumn{13}{c}{Baselines} \\\midrule
BART Canonical    & 41.43 & 58.57 & 68.29 & 23.21 & 35.71 & 63.10 & 29.81 & 42.86 & 55.90 & 20.18 & 48.19 & 55.12 \\ \midrule
 \multicolumn{13}{c}{Our models} \\\midrule
JT & 48.85 & 63.17 & 68.29 & 51.19 & \bf 58.33 & 62.50 & 52.80 & 49.69 & 59.63 & 58.43 & 64.46 & 66.57 \\
JT + ST & 56.52 & \bf 66.75 & \bf 71.61 & \bf 56.55 & 57.74 & 66.67 & \bf 60.87 & \bf 57.76 & 62.11 & \bf 65.96 & \bf 71.99 & 65.36  \\
JT + ST + Paraphrasing & \bf 58.06 & 66.24 & 70.84 & 55.95 & 55.95 & \bf 67.26 & 59.63 & 55.90 & \bf 62.73 &  65.66 & 69.58 & \bf 69.28\\\midrule
 \multicolumn{13}{c}{Reference} \\\midrule
Full BART Canonical &  \multicolumn{3}{c}{89.51 (n = 1561)} &  \multicolumn{3}{c}{ 85.12 (n = 669)}  &  \multicolumn{3}{c}{84.72 (n = 864)}  & \multicolumn{3}{c}{82.18 (n = 3535)} \\ \midrule
& \multicolumn{3}{c}{Blocks}& \multicolumn{3}{c}{Housing}& \multicolumn{3}{c}{Recipes}& \multicolumn{3}{c}{Social}
\\\cmidrule(lr){2-4}\cmidrule(lr){5-7}\cmidrule(lr){8-10}\cmidrule(lr){11-13}
           & n=16  & n=32 & n=48 & n=16  & n=32 & n=48 & n=16  & n=32 & n=48 & n=16  & n=32 & n=48 \\\midrule
          \multicolumn{13}{c}{Baselines} \\\midrule
BART Canonical    & 27.07 & 21.05 & 30.33 & 15.87 & 43.92 & 37.57 & 28.70 & 38.43 & 46.76 &  28.73 & 40.50 & 45.02\\
\midrule
 \multicolumn{13}{c}{Our models} \\\midrule
JT & 36.34 & 36.84 & 48.37 & 45.50 & 58.73 & 60.32 & 48.61 & \bf 60.19 & 63.89 & 24.21 & 39.48 & 53.17 \\
JT + ST & 38.85 & 38.60 & \bf 51.38 & \bf 52.91 & 64.02 & \bf 61.38 & 52.78 & 58.80 & 65.74 & \bf 36.09 & \bf 48.30 & 57.01 \\
JT + ST + Paraphrasing & \bf 39.35 & \bf 39.60 & 49.87 & 52.38 & \bf 64.55 & \bf 61.38 & \bf  53.70 & \bf 60.19 & \bf 66.20 &  31.45 & 45.25 & \bf 58.14\\\midrule
 \multicolumn{13}{c}{Reference} \\\midrule
Full BART Canonical &  \multicolumn{3}{c}{69.67 (n = 1596)} &  \multicolumn{3}{c}{80.42 (n = 752)}  &  \multicolumn{3}{c}{83.85 (n = 640)}  & \multicolumn{3}{c}{88.25 (n = 1325)} \\
\bottomrule
\end{tabular}}
\caption{Results on the Overnight dataset.
}
\label{tab:on}
\end{table*}

\subsection{Metrics}
We report the recommended variants of exact match (EM) accuracy for both Pizza and Overnight datasets. 

\textit{Unordered Exact Match Accuracy}:
For Pizza, we report unordered EM accuracy. This accounts for 
parses which have identical semantics but vary in their linearized representations due to differences
in sibling order. 

\textit{Denotation Accuracy}:
For Overnight, we report denotation accuracy. We execute the golden and predicted 
queries on the database and check for an exact match on the results. This accounts for any 
surface-level differences in the database queries that disappear upon actual execution.

\subsection{Model Details}

We use BART-Large as our base architecture. It contains 12 transformer encoder and decoder layers, 
16 attention heads, and embeddings of size 1024 ($\sim 458$ million parameters). 

We train all our models with sequence cross entropy loss using the Adam optimizer with
$\beta_1 = 0.9$, $\beta_2 = 0.98$, $\epsilon = 1e-9$ and the Noam LR scheduler  \citep{vaswani17} with 
500 warmup steps and a learning rate scale factor of 0.15. JT models are trained for 10 epochs, while base models
are trained for 100 to 1000 epochs on the low-resource data. We fix the batch size to 512 tokens for all models. 
We use dropout of 0.1 and freeze the encoder token and position embeddings during training. During inference, we use beam
search decoding with beam size 4. We did not perform any explicit hyperparameter tuning. 
Additional details, including the pizza canonicalization scheme, are provided in the appendix on our project page\footnote{https://github.com/amazon-research/resource-constrained-naturalized-semantic-parsing}, along with our data files.

\begin{table*}[t!]
\centering
{\footnotesize 
\begin{tabular}{lcccccccc}\toprule
& Basketball & Blocks & Calendar & Housing & Publications & Recipes & Restaurants & Social
\\\midrule
 \multicolumn{9}{c}{Baselines} \\\midrule
BART Canonical (our version) & 84.7 & 55.4 & 77.4 & 68.3 & 73.3 & 75.5 & 75.3 & 69.7\\
BART Canonical \citep{shin2021constrained} & 86.4 & 55.4 & 78.0 & 67.2 & 75.8 & 80.1 & 80.1 & 66.6 \\
GPT-3 \citep{shin2021constrained} & 85.9 & \bf 63.4 & 79.2 & \bf 74.1 & 77.6 & 79.2 & \bf 84.0 & 68.7 \\
\citet{cao2019semantic} & 77.2 & 42.9 & 61.3 & 55.0 & 69.6 & 67.1 & 63.9 & 56.6 \\\midrule
 \multicolumn{9}{c}{Our models} \\\midrule
JT & 84.9 & 62.7 & 81.0 & \bf 74.1 & 78.3 & 79.6 & 79.8 & 69.0 \\
JT + ST & \bf 87.7 & 62.4 & 82.1 & \bf 74.1 & 79.5 & \bf 81.9 & 80.7 & 69.2 \\
JT + ST + Paraphrasing & 86.7 & \bf 63.4 & \bf 83.3 & 73.5 & \bf 80.1 & 78.7 & 81.0 & \bf 69.9\\\bottomrule
\end{tabular}}
\caption{Results on the Overnight dataset with 200 training examples.}
\label{tab:on2}
\end{table*}

\section{Results}
\label{sec:res}

Table~\ref{tab:pizza} shows the performance of our models and baselines on Pizza. 
A fully trained BART model with canonical-form targets and constrained decoding (trained
on the full dataset of 348 examples) achieves an unordered EM accuracy of 87.25\%. This is the SOTA result on this 
dataset. However, when the training data is reduced to 16 examples, the score drops to 16.95\%. 
We see similar significant drops with 32 and 48 examples, with scores of 53.35\% and 58.36\% respectively. 
JT gives a huge boost to all three settings. With 16 examples, the accuracy 
jumps to 42.23\%, with 32 examples to 64.70\%, and with 48 examples to 70.30\%. Self-training and
paraphrase augmentation  provide further boosts in the 16 and 48 example settings. Overall, we see
that our best scores greatly improve upon the performance of the base architecture. This effect
is most apparent in the 16 example setting, where we obtain an almost $3\times$ improvement.

We see similar result trends with the Overnight dataset. Table~\ref{tab:on} shows the denotation accuracies
of all models across all eight domains. Our JT models attain a significant improvement over the baselines and bridge
the gap towards fully trained models across all domains. The trend is especially noticeable in the calendar and
recipes domains in the 16 example setting, where the denotation accuracies jump from 23.21\% and 28.70\% to 56.55\%
and 53.70\% for our best models, respectively, a $2\times$ boost on average.

We wanted to analyze how far these improvements hold, so we repeated
the experiment with 200 examples on the Overnight dataset. This also allows us to make a more direct comparison
with some prior works that reported results for this setting. 
Table~\ref{tab:on2} presents these results. We see that our proposed techniques improve
the denotation accuracies across all the domains over our baseline BART model by roughly 5 absolute points on average. Overall, they even slightly outperform
 the much larger and access-restricted GPT-3 model reported by \citet{shin2021constrained}.

\section{Analysis}
We analyzed some of our design decisions with experiments on the Pizza dataset. 
A detailed analysis can be found in the appendix on our project page listed as a footnote in the previous page.  
We summarize some of our findings here.

\textit{Two-stage Finetuning}: Our JT approach, while being simpler, does at-least as well or better than a two-stage finetuning process, 
where the auxiliary tasks are first used to pretrain the model and then the annotated data is used to finetune it. 
We see a noticeable drop in accuracy with the extra fine-tuning step for 32 ($65\% \rightarrow 59\%$) and 48 ($70\% \rightarrow 67\%$) examples,
and no significant boost for 16 examples. 

\textit{Importance of the canonical form}: For our JT technique, the canonical
form provides us with an easy way to add meaningful noise without modifying the content tokens for the 
denoising auxiliary task. We can simply perform token level operations without worrying about the target
structure. If the targets are parse trees, adding noise is trickier, since most of the tokens 
in the parse represent content and meaningful operations that need to be performed at the tree level. Further,
the target sequences for the mask prediction task are in natural language and are better aligned with the canonical
form targets than the parse trees. This potentially allows for better knowledge transfer during joint training. 

We performed a JT experiment with a model that predicts tree LFs instead of canonical forms.
We created the source sequences for the denoising auxiliary task 
using tree-level noise operations such as switching entities, dropping
brackets, and inserting random tokens. We found that the resulting models achieved significantly lower scores than the models
that use canonical targets. For the 48 example case, the LF model achieves 59\% accuracy compared to our JT model's 70\%.

\textit{Synthetic data auxiliary task}: The goal of our auxiliary tasks was to provide the model with a challenging objective. 
To train the decoder, we use the synthetically generated target sequences so that the decoder can train on,
and learn to generate, a variety of valid canonical forms.  To create a challenge for the decoder, we noise
the targets to obtain corrupted source sequences and create a denoising task. 

However, there are other possible tasks. One could create rules to generate synthetic utterances 
given the target parses. This synthetic data could then be used to train the decoder. This approach, however, requires 
manual effort and depends on the quality and diversity of the synthetic data. For Pizza, we
already have access to synthetic data, since the entire training set is synthetic. Assuming we
have access to a system that can generate such synthetic utterances given randomly generated target parses,
we could replace our denoising task with the synthetic examples. We perform 
this experiment to compare these two auxiliary tasks.  The synthetic model achieves 82\% accuracy compared our denoising 
model's 70\% in the 48 example case.
The synthetic parsing auxiliary task performs better than denoising but requires lots of manual effort to create a synthetic utterance grammar. Our JT approach is directly applicable to both tasks.

\section{Related Work}

Naturalized semantic parsing can be traced back to work by \citet{BerantLiang2014}, who introduced the idea
of canonical natural-language formulations of utterances. Our base architecture is based on work 
by \citet{shin2021constrained}. There have been other approaches that explored low-resource semantic parsing in the past, which
used concepts from meta-learning, self-training, and synthetic data generation \citep{goldwasser2011confidence,xu2020autoqa,mcclosky2006effective}. 
Our model, however, is designed to be applicable to extremely small data sizes without requiring any external manual effort.

\citet{wu2021paraphrasing} have explored unsupervised semantic parsing as paraphrasing by decoding controlled paraphrases using a synchronous grammar. Accordingly, that approach requires a carefully crafted synchronous grammar, whereas our method relies on readily available data for auxiliary tasks.

Recently, there has also been an upward trend towards in-context learning or ``prompting'' approaches in low-resource
settings \citep{brown2020language,shin2021constrained}.
In these approaches, massive LMs are directly used to solve tasks without any training by framing the 
task as a prompt in the style of the pretraining objective, with a few task demonstrations selected from the handful of annotated examples.
However, only GPT-3 has been shown to work well with a generation-style parsing task; smaller architectures, such as GPT-2, 
could not replicate the performance \citep{shin2021constrained}. GPT-3 is a 175-billion parameter model that
is currently not accessible to the entire research community.

Our JT technique can be seen as a mixture of domain adaptation of the pretrained LM and a regularizer.
GRAPPA \citep{yu2020grappa} is a recent effort that improves table semantic parsing using a separate pretraining phase, where 
the model is trained on synthetic parsing data and table-related utterances for domain adaptation before fine-tuning on 
a small annotated dataset of around 10k examples. Our work is similar to GRAPPA but focuses on much smaller training datasets, which 
requires us to train our model jointly with auxiliary tasks to make it more robust. 
We also show that denoising canonical forms is a reasonable auxiliary task.

At a high level, our approach also has some similarities to the work of \citet{schickSchutze2021NotJustSize},
who also aim to show that smaller---and greener---LMs can be effective few-shot learners. They also utilize unlabeled data and a form of
self-learning to augment a small amount of golden annotations. However, they focus on classification rather than generation tasks
(reducing classification tasks to MLM).

Constraining the decoder of a neural semantic parser so that beam search only considers paths that adhere to various syntactic or semantic
constraints has been widely explored over the last few years \citep{krishnamurthy-etal-2017-neural,YinN17}.  \citet{XiaoConstraining} show that constrained decoding can result in significant latency improvements. 

\section{Conclusions}
Our key idea is the application of joint training with auxiliary tasks to train low-resource semantic parsing models. The data for the auxiliary tasks is constructed from  unlabeled data and combined with the limited annotated data during training. We also introduce self-training and paraphrasing steps to augment the initial data and further improve model performance.

We start with a strong baseline architecture that uses a BART-Large model, canonical-form targets, and constrained decoding,
and show that our techniques provide massive improvements, in the order of $2$--$3\times$ on EM scores. We evaluate our
models on two datasets, Pizza and Overnight (the latter containing eight separate domains), and on three data sizes: 16, 32, and 48 examples.
Models trained with our techniques consistently show improvements over baseline 
architectures across all datasets and size settings. The improvements are especially notable in the scarcest setting with
16 annotated examples. We analyze our model design and results in another series of experiments and show the effectiveness
of our approach in constructing a robust, well-performing semantic parsing model. 

\bibliographystyle{ijcai22}
\bibliography{ijcai22}

\begin{thebibliography}{}

\bibitem[\protect\citeauthoryear{Berant and Liang}{2014}]{BerantLiang2014}
Jonathan Berant and Percy Liang.
\newblock Semantic parsing via paraphrasing.
\newblock In {\em ACL (1)}, pages 1415--1425. ACL (Association for Computer
  Linguistics), 2014.

\bibitem[\protect\citeauthoryear{Berant \bgroup \em et al.\egroup
  }{2013}]{berant2013freebase}
J.~Berant, A.~Chou, R.~Frostig, and P.~Liang.
\newblock Semantic parsing on {F}reebase from question-answer pairs.
\newblock In {\em Empirical Methods in Natural Language Processing (EMNLP)},
  2013.

\bibitem[\protect\citeauthoryear{Brown \bgroup \em et al.\egroup
  }{2020}]{brown2020language}
Tom~B Brown, Benjamin Mann, Nick Ryder, Melanie Subbiah, Jared Kaplan, Prafulla
  Dhariwal, Arvind Neelakantan, Pranav Shyam, Girish Sastry, Amanda Askell,
  et~al.
\newblock Language models are few-shot learners.
\newblock {\em arXiv preprint arXiv:2005.14165}, 2020.

\bibitem[\protect\citeauthoryear{Cao \bgroup \em et al.\egroup
  }{2019}]{cao2019semantic}
Ruisheng Cao, Su~Zhu, Chen Liu, Jieyu Li, and Kai Yu.
\newblock Semantic parsing with dual learning.
\newblock {\em arXiv preprint arXiv:1907.05343}, 2019.

\bibitem[\protect\citeauthoryear{Dong \bgroup \em et al.\egroup
  }{2018}]{NSPConfidence}
Li~Dong, Chris Quirk, and Mirella Lapata.
\newblock Confidence modeling for neural semantic parsing.
\newblock In {\em Proceedings of the 56th Annual Meeting of the Association for
  Computational Linguistics (Volume 1: Long Papers)}, pages 743--753,
  Melbourne, Australia, July 2018. Association for Computational Linguistics.

\bibitem[\protect\citeauthoryear{Einolghozati \bgroup \em et al.\egroup
  }{2019}]{einolghozati2019improving}
Arash Einolghozati, Panupong Pasupat, Sonal Gupta, Rushin Shah, Mrinal Mohit,
  Mike Lewis, and Luke Zettlemoyer.
\newblock Improving semantic parsing for task oriented dialog.
\newblock {\em arXiv preprint arXiv:1902.06000}, 2019.

\bibitem[\protect\citeauthoryear{Feng \bgroup \em et al.\egroup
  }{2021}]{feng2021survey}
Steven~Y Feng, Varun Gangal, Jason Wei, Sarath Chandar, Soroush Vosoughi,
  Teruko Mitamura, and Eduard Hovy.
\newblock A survey of data augmentation approaches for nlp.
\newblock {\em arXiv preprint arXiv:2105.03075}, 2021.

\bibitem[\protect\citeauthoryear{Goldwasser \bgroup \em et al.\egroup
  }{2011}]{goldwasser2011confidence}
Dan Goldwasser, Roi Reichart, James Clarke, and Dan Roth.
\newblock Confidence driven unsupervised semantic parsing.
\newblock In {\em Proceedings of the 49th Annual Meeting of the Association for
  Computational Linguistics: Human Language Technologies}, pages 1486--1495,
  2011.

\bibitem[\protect\citeauthoryear{Krishnamurthy \bgroup \em et al.\egroup
  }{2017}]{krishnamurthy-etal-2017-neural}
Jayant Krishnamurthy, Pradeep Dasigi, and Matt Gardner.
\newblock Neural semantic parsing with type constraints for semi-structured
  tables.
\newblock In {\em Proceedings of the 2017 Conference on Empirical Methods in
  Natural Language Processing}, pages 1516--1526, Copenhagen, Denmark,
  September 2017. Association for Computational Linguistics.

\bibitem[\protect\citeauthoryear{Lewis \bgroup \em et al.\egroup
  }{2019}]{lewis2019bart}
Mike Lewis, Yinhan Liu, Naman Goyal, Marjan Ghazvininejad, Abdelrahman Mohamed,
  Omer Levy, Ves Stoyanov, and Luke Zettlemoyer.
\newblock Bart: Denoising sequence-to-sequence pre-training for natural
  language generation, translation, and comprehension.
\newblock {\em arXiv preprint arXiv:1910.13461}, 2019.

\bibitem[\protect\citeauthoryear{McClosky \bgroup \em et al.\egroup
  }{2006}]{mcclosky2006effective}
David McClosky, Eugene Charniak, and Mark Johnson.
\newblock Effective self-training for parsing.
\newblock In {\em Proceedings of the Human Language Technology Conference of
  the NAACL, Main Conference}, pages 152--159, 2006.

\bibitem[\protect\citeauthoryear{Mihalcea}{2004}]{mihalcea2004co}
Rada Mihalcea.
\newblock Co-training and self-training for word sense disambiguation.
\newblock In {\em Proceedings of the Eighth Conference on Computational Natural
  Language Learning (CoNLL-2004) at HLT-NAACL 2004}, pages 33--40, 2004.

\bibitem[\protect\citeauthoryear{Rongali \bgroup \em et al.\egroup
  }{2020}]{rongali2020don}
Subendhu Rongali, Luca Soldaini, Emilio Monti, and Wael Hamza.
\newblock Don’t parse, generate! a sequence to sequence architecture for
  task-oriented semantic parsing.
\newblock In {\em Proceedings of The Web Conference 2020}, pages 2962--2968,
  2020.

\bibitem[\protect\citeauthoryear{Schick and
  Sch{\"u}tze}{2021}]{schickSchutze2021NotJustSize}
Timo Schick and Hinrich Sch{\"u}tze.
\newblock It{'}s not just size that matters: Small language models are also
  few-shot learners.
\newblock In {\em Proceedings of the 2021 Conference of the North American
  Chapter of the Association for Computational Linguistics: Human Language
  Technologies}, pages 2339--2352. Association for Computational Linguistics,
  June 2021.

\bibitem[\protect\citeauthoryear{Shin \bgroup \em et al.\egroup
  }{2021}]{shin2021constrained}
Richard Shin, Christopher~H Lin, Sam Thomson, Charles Chen, Subhro Roy,
  Emmanouil~Antonios Platanios, Adam Pauls, Dan Klein, Jason Eisner, and
  Benjamin Van~Durme.
\newblock Constrained language models yield few-shot semantic parsers.
\newblock {\em arXiv preprint arXiv:2104.08768}, 2021.

\bibitem[\protect\citeauthoryear{Vaswani \bgroup \em et al.\egroup
  }{2017}]{vaswani17}
Ashish Vaswani, Noam Shazeer, Niki Parmar, Jakob Uszkoreit, Llion Jones, Aidan
  N.~Gomez, Lukasz Kaiser, and Illia Polosukhin.
\newblock Attention is all you need.
\newblock 2017.

\bibitem[\protect\citeauthoryear{Wang \bgroup \em et al.\egroup
  }{2015}]{wang2015building}
Yushi Wang, Jonathan Berant, and Percy Liang.
\newblock Building a semantic parser overnight.
\newblock In {\em Proceedings of the 53rd Annual Meeting of the Association for
  Computational Linguistics and the 7th International Joint Conference on
  Natural Language Processing (Volume 1: Long Papers)}, pages 1332--1342, 2015.

\bibitem[\protect\citeauthoryear{Wieting and Gimpel}{2017}]{wieting2017paranmt}
John Wieting and Kevin Gimpel.
\newblock Paranmt-50m: Pushing the limits of paraphrastic sentence embeddings
  with millions of machine translations.
\newblock {\em arXiv preprint arXiv:1711.05732}, 2017.

\bibitem[\protect\citeauthoryear{Wu \bgroup \em et al.\egroup
  }{2021}]{wu2021paraphrasing}
Shan Wu, Bo~Chen, Chunlei Xin, Xianpei Han, Le~Sun, Weipeng Zhang, Jiansong
  Chen, Fan Yang, and Xunliang Cai.
\newblock From paraphrasing to semantic parsing: unsupervised semantic parsing
  via synchronous semantic decoding.
\newblock {\em arXiv preprint arXiv:2106.06228}, 2021.

\bibitem[\protect\citeauthoryear{Xiao \bgroup \em et al.\egroup
  }{2019}]{XiaoConstraining}
Chunyang Xiao, Christoph Teichmann, and Konstantine Arkoudas.
\newblock Grammatical sequence prediction for real-time neural semantic
  parsing.
\newblock In {\em Proceedings of the Workshop on Deep Learning and Formal
  Languages: Building Bridges}, pages 14--23, Florence, August 2019. ACL
  (Association for Computational Linguistics).

\bibitem[\protect\citeauthoryear{Xu \bgroup \em et al.\egroup
  }{2020}]{xu2020autoqa}
Silei Xu, Sina~J Semnani, Giovanni Campagna, and Monica~S Lam.
\newblock Autoqa: From databases to qa semantic parsers with only synthetic
  training data.
\newblock {\em arXiv preprint arXiv:2010.04806}, 2020.

\bibitem[\protect\citeauthoryear{Yin and Neubig}{2017}]{YinN17}
Pengcheng Yin and Graham Neubig.
\newblock A syntactic neural model for general-purpose code generation.
\newblock In Regina Barzilay and Min{-}Yen Kan, editors, {\em Proceedings of
  the 55th Annual Meeting of the Association for Computational Linguistics,
  {ACL} 2017, Vancouver, Canada, July 30 - August 4, Volume 1: Long Papers},
  pages 440--450. Association for Computational Linguistics, 2017.

\bibitem[\protect\citeauthoryear{Yu \bgroup \em et al.\egroup
  }{2020}]{yu2020grappa}
Tao Yu, Chien-Sheng Wu, Xi~Victoria Lin, Bailin Wang, Yi~Chern Tan, Xinyi Yang,
  Dragomir Radev, Richard Socher, and Caiming Xiong.
\newblock Grappa: Grammar-augmented pre-training for table semantic parsing.
\newblock {\em arXiv preprint arXiv:2009.13845}, 2020.

\end{thebibliography}

\end{document}